%% file: paper947.tex
\documentclass[runningheads]{llncs}
\usepackage{graphicx}
\usepackage{amsmath}
\usepackage{amssymb}
\usepackage{booktabs}
\usepackage{rotating}
\usepackage{makecell}
\usepackage{xspace}
\usepackage{xcolor}
\usepackage[export]{adjustbox}
\usepackage{multirow}
\usepackage{float}
\usepackage{pifont}
\usepackage{caption}
\usepackage{subcaption}
\usepackage{multirow}
\usepackage[pagebackref,breaklinks,colorlinks]{hyperref}

\usepackage[capitalize]{cleveref}
\crefname{section}{Sec.}{Secs.}
\Crefname{section}{Section}{Sections}
\Crefname{table}{Table}{Tables}
\crefname{table}{Tab.}{Tabs.}

\makeatletter
\DeclareRobustCommand\onedot{\futurelet\@let@token\@onedot}
\def\@onedot{\ifx\@let@token.\else.\null\fi\xspace}
 
\def\eg{\emph{e.g}\onedot} 
\def\ie{\emph{i.e}\onedot}

\def\etal{\emph{et al}\onedot}
\makeatother

\newcommand*\samethanks[1][\value{footnote}]{\footnotemark[#1]}

\begin{document}
\title{EchoGLAD: Hierarchical Graph Neural Networks for Left Ventricle Landmark Detection on Echocardiograms}
\titlerunning{Graph Neural Networks for Left Ventricle Landmark Detection}
\author{Masoud Mokhtari\inst{1}\orcidID{0000-0001-9471-5573} \and
Mobina Mahdavi \inst{1} \and
Hooman Vaseli \inst{1}\orcidID{0000-0002-8259-9488} \and
Christina Luong \inst{2} \and
Purang Abolmaesumi\inst{1}\thanks{Co-Corresponding Authors}\and
Teresa S. M. Tsang\inst{2} \and
Renjie Liao\inst{1}\samethanks}
% index{Mokhtari, Masoud}
% index{Mahdavi, Mobina}
% index{Vaseli, Hooman}
% index{Luong, Christina}
% index{Abolmaesumi, Purang}
% index{Tsang, Teresa}
% index{Liao, Renjie}
%
\authorrunning{M. Mokhtari et al.}
\institute{Electrical and Computer Engineering, University of British Columbia,
Vancouver, BC, Canada \\
\email{\{masoud, mobina, hoomanv, purang, rjliao\}@ece.ubc.ca} \and
Vancouver General Hospital, Vancouver, BC, Canada\\
\email{\{christina.luong, t.tsang\}@ubc.ca}}
\maketitle              % typeset the header of the contribution
\begin{abstract}
The functional assessment of the left ventricle chamber of the heart requires detecting four landmark locations and measuring the internal dimension of the left ventricle and the approximate mass of the surrounding muscle.
The key challenge of automating this task with machine learning is the sparsity of clinical labels, i.e., only a few landmark pixels in a high-dimensional image are annotated, leading many prior works to heavily rely on isotropic label smoothing. 
However, such a label smoothing strategy ignores the anatomical information of the image and induces some bias.
To address this challenge, we introduce an \textbf{echo}cardiogram-based, hierarchical \textbf{g}raph neural network (GNN) for left ventricle \textbf{la}ndmark \textbf{d}etection (EchoGLAD). 
Our main contributions are: 1) a hierarchical graph representation learning framework for multi-resolution landmark detection via GNNs; 2) induced hierarchical supervision at different levels of granularity using a multi-level loss. 
We evaluate our model on a public and a private dataset under the in-distribution (ID) and out-of-distribution (OOD) settings. 
For the ID setting, we achieve the state-of-the-art mean absolute errors (MAEs) of 1.46~mm and 1.86~mm on the two datasets. 
Our model also shows better OOD generalization than prior works with a testing MAE of 4.3~mm. 
    
\end{abstract}

\section{Introduction}
\label{sec:intro}

\begin{figure}
    \centering 
    \begin{tabular}{@{\hspace{0mm}}c@{\hspace{0.5mm}}c@{\hspace{1mm}}c@{\hspace{1mm}}c@{\hspace{1mm}}c@{\hspace{1mm}}c@{\hspace{1mm}}c@{\hspace{1mm}}c@{\hspace{1mm}}c}
        \includegraphics[width=0.35\textwidth,trim={0 8 0 0},clip,valign=m]{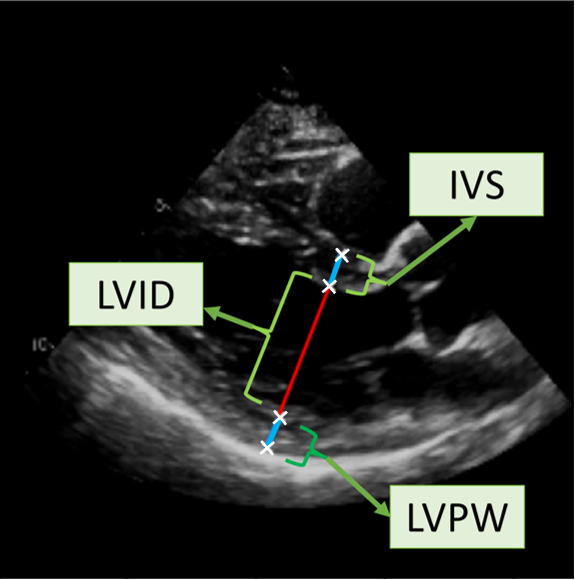} & 
        \includegraphics[width=0.35\textwidth,trim={0 0 0 0},clip,valign=m]{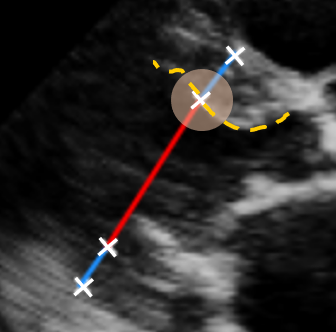} &            
        \\ 
        {\footnotesize  } 
        {\footnotesize (a) LV Measurements} 
        & {\footnotesize (b) Label Smoothing}
        \\
    \end{tabular}
    \caption{(a) IVS, LVID and LVPW measurements visualized on a PLAX echo frame. (b) If the wall landmark labels are smoothed by an isotropic Gaussian distribution, points along the visualized wall and ones perpendicular are penalized equally. Ideally, points along the walls must be penalized less.} 
    \label{fig: plax}
\end{figure}

Left Ventricular Hypertrophy (LVH), one of the leading predictors of adverse cardiovascular outcomes, is the condition where heart’s mass abnormally increases secondary to anatomical changes in the Left Ventricle (LV)~\cite{GRADMAN2006326}. These anatomical changes include an increase in the septal and LV wall thickness, and the enlargement of the LV chamber. More specifically, Inter-Ventricular Septal (IVS), LV Posterior Wall (LVPW) and LV Internal Diameter (LVID) are assessed to investigate LVH and the risk of heart failure~\cite{McFarland1978}. As shown in Figure \ref{fig: plax} (a), four landmarks on a parasternal long axis (PLAX) echo frame can characterize IVS, LVPW and LVID, and allow cardiac function assessment. To automate this, machine learning-based (ML) landmark detection methods have gained traction. 

It is difficult for such ML models to achieve high accuracy due to the sparsity of positive training signals (four or six) pertaining to the correct pixel locations. In an attempt to address this, previous works use 2D Gaussian distributions to smooth the ground truth landmarks of the LV~\cite{jamie,jafari2021u,lin2021reciprocal}. However, as shown in Figure \ref{fig: plax} (b), for LV landmark detection where landmarks are located at the wall boundaries (as illustrated by the dashed line), we argue that an isotropic Gaussian label smoothing approach confuses the model by being agnostic to the structural information of the echo frame and penalizing the model similarly whether the predictions are perpendicular or along the LV walls.

\begin{figure}
    \centering
    \includegraphics[width=0.99\linewidth]{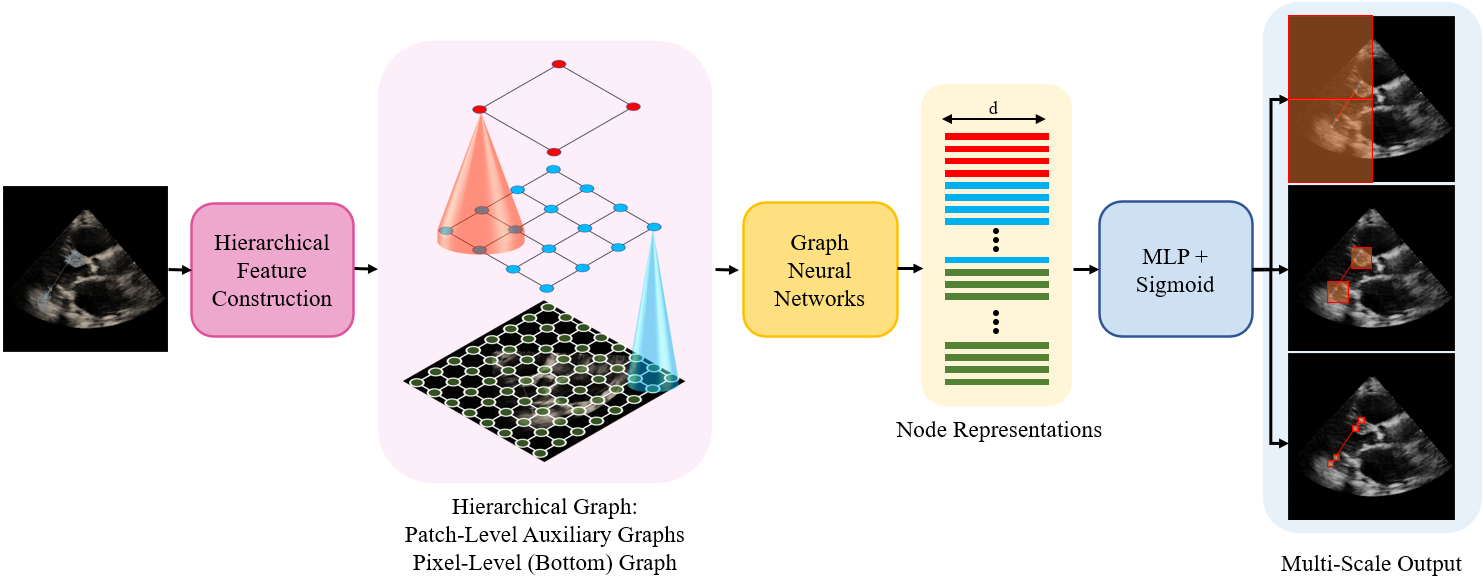}
    \caption{Overview of our proposed model architecture. \textbf{Hierarchical Feature Construction} provides node features for the hierarchical graph representation of each echo frame where the nodes in the main graph correspond to pixels in the image, and nodes in the auxiliary graphs correspond to patches of different granularity in the image. \textbf{Graph Neural Networks} are used to process the hierarchical graph representation and produce node embeddings for the auxiliary graphs and the main graph. \textbf{Multi-Layer Perceptrons (MLPs)} are followed by a Sigmoid output function to map the node embeddings into landmark heatmaps of different granularity over the input echo frame.}    
    \label{fig: overall_arch}
\end{figure}

In this work, to address the challenge brought by sparse annotations and label smoothing, we propose a hierarchical framework based on Graph Neural Networks (GNNs) \cite{scarselli2008graph} to detect LV landmarks in ultrasound images. 
As shown in Figure~\ref{fig: overall_arch}, our framework learns useful representations on a hierarchical grid graph built from the input echo image and performs multi-level prediction tasks. 

Our contributions are summarized below.
\begin{itemize}
    \item[$\bullet$] We propose a novel GNN framework for LV landmark detection, performing message passing over hierarchical graphs constructed from an input echo;
    \item[$\bullet$] We introduce a hierarchical supervision that is automatically induced from sparse annotations to alleviate the issue of label smoothing;
    \item[$\bullet$] We evaluate our model on two LV landmark datasets and show that it not only achieves state-of-the-art mean absolute errors (MAEs) (1.46 mm and 1.86 mm across three LV measurements) but also outperforms other methods in out-of-distribution (OOD) testing (achieving 4.3~mm).    
\end{itemize}

\section{Related Work}

Various convolution-based LV landmark detection works have been proposed. Sofka \etal~\cite{sofka2017fully} use Fully Convolutional Networks to generate prediction heatmaps followed by a center of mass layer to produce the coordinates of the landmark locations. Another work~\cite{lin2021reciprocal} uses a modified U-Net \cite{unet} model to produce a segmentation map followed by a focal loss to penalize pixel predictions in close proximity of the ground truth landmark locations modulated by a Gaussian distribution. Jafari \etal~\cite{jafari2021u} use a similar U-Net model with Bayesian neural networks~\cite{bnn} to estimate the uncertainty in model predictions and reject samples that exhibit high uncertainties. Gilbert \etal~\cite{gilbert2019automated} smooth ground truth labels by placing 2D Gaussian heatmaps around landmark locations at angles that are statistically obtained from training data. Lastly, Duffy \etal~\cite{echonetlvh} use atrous convolutions \cite{atrous} to make predictions for LVID, IVS and LVPW measurements.

Other related works focus on the detection of cephalometric landmarks from X-ray images. These works are highly transferable to the task of LV landmark detection as they must also detect a sparse number of landmarks. McCouat \etal~\cite{contour} is one of these works that abstains from using Gaussian label smoothing, but still relies on one-hot labels and treats landmark detection as a pixel-wise classification task. Chen \etal~\cite{pyramid} is another cephalometric landmark detection work that creates a feature pyramid from the intermediate layers of a ResNet~\cite{resnet}.

Our approach is different from prior works in that it aims to avoid the issue shown in \cref{fig: plax} (b) and the sparse annotations problem by the introduction of simpler auxiliary tasks to guide the main pixel-level task, so that the ML model learns the location of the landmarks without relying on Gaussian label smoothing. 
It further improves the representation learning via efficient message-passing~\cite{scarselli2008graph,messagepassing} of GNNs among pixels and patches at different levels without having as high a computational complexity as transformers~\cite{vision_transformer,swin}.
Lastly, while GNNs have never been applied to the task of LV landmark detection, they have been used for landmark detection in other domains. Li \etal~\cite{li2020structured} and Lin \etal~\cite{lin2021structure} perform face landmark detection via modeling the landmarks with a graph and performing a cascaded regression of the locations. 
These methods, however, do not leverage hierarchical graphs and hierarchical supervision and instead rely on initial average landmark locations, which is not an applicable approach to echo, where the anatomy of the depicted heart can vary significantly. Additionally, Mokhtari \etal~\cite{mokhtari2022} use GNNs for the task of EF prediction from echo cine series. However, their work focuses on regression tasks.

\section{Method}
\label{sec: method}
\subsection{Problem Setup}
We consider the following supervised setting for LV wall landmark detection. We have a dataset $D = \{X, Y\}$, where $|D| = n$ is the number of $\{x^i, y^i\}$ pairs such that $x^i \in X$, $y^i \in Y$, and $i \in [1, n]$. Each $x^i \in \mathbb{R}^{H\times W}$ is an echo image of the heart, where H and W are height and width of the image, respectively, and each $y^i$ is the set of four point coordinates $[(h^i_1, w^i_1), (h^i_2, w^i_2), (h^i_3, w^i_3), (h^i_4, w^i_4)]$ indicating the landmark locations in $x^i$. Our goal is to learn a function ${f}: \mathbb{R}^{H \times W} \mapsto 
\mathbb{R}^{4 \times 2}$ that predicts the four landmark coordinates for each input image. \textit{A figure in the supp. material further clarifies how the model generates landmark location heatmaps on different scales (Fig. 2).}

\subsection{Model Overview}
\label{sec: model_arch}
As shown in Figure~\ref{fig: overall_arch}, each input echo frame is represented by a hierarchical grid graph where each sub-graph corresponds to the input echo frame at a different resolution. The model produces heatmaps over both the main pixel-level task as well as the coarse auxiliary tasks. While the pixel-level heatmap prediction is of main interest, we use a hierarchical multi-level loss approach where the model's prediction over auxiliary tasks is used during training to optimize the model through comparisons to coarser versions of the ground truth. The intuition behind such an approach is that the model learns nuances in the data by performing landmark detection on the easier auxiliary tasks and uses this established reasoning when performing the difficult pixel-level task.

\subsection{Hierarchical Graph Construction}

\label{sec: graph_creation}
To learn representations that better capture the dependencies among pixels and patches, we introduce a hierarchical grid graph along with multi-level prediction tasks. 
As an example, the simplest task consists of a grid graph with only four nodes, where each node corresponds to four equally-sized patches in the original echo image. 
In the main task (the one that is at the bottom in Figure \ref{fig: overall_arch} and is the most difficult), the number of nodes is equal to the total number of pixels.

More formally, let us denote a graph as $G = (V, E)$, where $V$ is the set of nodes, and $E$ is the set of edges in the graph such that if $v_i, v_j \in V$ and there is an edge from $v_i$ to $v_j$, then $e_{i,j} \in E$.
To build hierarchical task representations, for each image $x \in X$ and the ground truth $y \in Y$, $K$ different auxiliary graphs $G_k(V_k, E_k)$ are constructed using the following steps for each $k \in [1,K]$:
\begin{enumerate}
    \item $2^k \times 2^k = 4^k$ nodes are added to $V_k$ to represent each patch in the image. Note that the larger values of $k$ correspond to graphs of finer resolution, while the smaller values of $k$ correspond to coarser graphs.
    \item Grid-like, undirected edges are added such that $e_{m-1, q}, e_{m+1, q}, e_{m, q-1}, e_{m, q+1} \in E_k$ for each $m, q \in [1 \dots 2^k]$ if these neighbouring nodes exist in the graph (border nodes will not have four neighbouring nodes).
    \item A patch feature embedding $z^k_j$, where $j \in [1 \dots 4^k]$ is generated and associated with that patch (node) $v_j \in V_k$. The patch feature construction technique is described in Section~\ref{sec: feat_gen}.
    \item Binary node labels $\hat{y}_k \in \{0, 1\}^{4^k \times 4}$ are generated such that $\hat{y}_{kj} = 1$ if at least one of the ground truth landmarks in $y$ is contained in the patch associated with node $v_j \in V_k$. Note that for each auxiliary graph, four different one-hot labels are predicted, which correspond to each of the four landmarks required to characterize LV measurements.
\end{enumerate}
The main graph, $G_\text{main}$, has a grid structure and contains $H \times W$ nodes regardless of the value of $K$, where each node corresponds to a pixel in the image. Additionally, to allow the model to propagate information across levels, we add inter-graph edges such that each node in a graph is connected to four nodes in the corresponding region in the next finer graph as depicted in \cref{fig: overall_arch}.

\subsection{Node Feature Construction}
\label{sec: feat_gen}
The graph representation described in Section~\ref{sec: graph_creation} is not complete without proper node features, denoted by $z \in \mathbb{R}^{|V|\times d}$, characterizing patches or pixels of the image. To achieve this, the grey-scale image is initially expanded in the channel dimension using a CNN. 
The features are then fed into a U-Net where the decoder part is used to obtain node features such that deeper layer embeddings correspond to the node features for the finer graphs.
This means that the main pixel-level graph would have the features of the last layer of the network. 
\textit{A figure clarifying node feature construction is provided in the supp. material (Fig. 1).}

\subsection{Hierarchical Message Passing}
\label{sec: intercom}
We now introduce how we perform message passing on our constructed hierarchical graph using GNNs to learn node representations for predicting landmarks. 

The whole hierarchical graph created for each sample, \ie, the main graph, auxiliary graphs, and cross-level edges, are collectively denoted as $G^i$, where $i\in [1, \dots, n]$. 
Each $G^i$ is fed into GNN layers followed by an MLP: 
\begin{align}
    h_{\text{nodes}}^{l+1} &= \text{ReLU}(\text{GNN}_l(G^i), h_{\text{nodes}}^l), \quad l \in [0, \dots, L] \\
    h_{\text{out}} &= \sigma(\text{MLP}(h_{\text{nodes}^{L+1}})) ,
\end{align}
where $\sigma$ is the Sigmoid function, $h_{\text{nodes}}^l \in \mathbb{R}^{|V_{G^i}|\times d}$ is the set of d-dimensional embeddings for all nodes in the graph at layer $l$, and $h_\text{out} \in [0,1]^{|V_{G^i}|\times 4}$ is the four-channel prediction for each node with each channel corresponding to a heatmap for each of the pixel landmarks.
The initial node features $h_{\text{nodes}}^1$ are set to the features $z$ described in Sections~\ref{sec: graph_creation} and \ref{sec: feat_gen}. The coordinates $(x_{\text{out}}^p, y_{\text{out}}^p)$ for each landmark location $p \in [1,2,3,4]$ are obtained by taking the expected value of individual heatmaps $h_\text{out}^p$ along the $x$ and $y$ directions such that:
\begin{align}
    x_{\text{out}}^p = \sum_{s=1}^{|V_{G^i}|} \text{softmax}(h_\text{out}^p)_s * \text{loc}_x(s) \label{eq: soft1},
\end{align}
where similar operations are performed in the y direction for $y_{\text{out}}^p$. Here, we vectorize the 2D heatmap into a single vector and then feed it to the softmax. 
$\text{loc}_x$ and $\text{loc}_y$ return the $x$ and $y$ positions of a node in the image. It must be noted that unlike some prior works such as Duffy \etal \cite{echonetlvh} that use post-processing steps such as imposing thresholds on the heatmap values, our work directly uses the output heatmaps to find the final predictions.

\subsection{Training and Objective Functions}\label{sec: objectives}
To train the network, we leverage two types of objective functions. 1) \emph{Weighted Binary Cross Entropy (BCE):} Since the number of landmark locations is much smaller than non-landmark locations, we use a weighted BCE loss; 2) 
 \emph{L2 regression of landmark coordinates:} We add a regression objective which is the L2 loss between the predicted coordinates and the ground truth labels.  

\section{Experiments}
\subsection{Datasets}
\label{sec: dataset}
\textbf{Internal Dataset:}
Our private dataset contains 29,867 PLAX echo frames, split in a patient-exclusive manner with 23824, 3004, and 3039 frames for training, validation, and testing, respectively.
\textbf{External Dataset:}
The public Unity Imaging Collaborative (UIC) \cite{uic} LV landmark dataset consists of a combination of 3822 end-systolic and end-diastolic PLAX echo frames acquired from seven British echocardiography labs. The provided splits contain 1613, 298, and 1911 training, validation, and testing samples, respectively. For both datasets, we down-sample the frames to a fixed size of $224 \times 224$.
\subsection{Implementation Details}
\label{sec: imp}
Our model creates $K$=7 auxiliary graphs. For the node features, the initial single-layer CNN uses a kernel size of 3 and zero-padding to output features with a dimension of $224\times224\times4$  ($C$=4). The U-Net's encoder contains $7$ layers with $128\times128, 64\times64, 32\times32, 16\times16, 8\times8, 4\times4$, and $2\times2$ spatial dimensions, and $8, 16, 32, 64, 128, 256$, and $512$ number of channels, respectively. Three Graph Convolutional Network (GCN)\cite{gcn} layers ($L=3$) with a hidden node dimension of 128 are used. To optimize the model, we use the Adam optimizer~\cite{adam} with an initial learning rate of 0.001, $\beta$ of (0.9, 0.999) and a weight decay of 0.0001, and for the weighted BCE loss, we use a weight of 9000. The model is implemented using PyTorch~\cite{pytorch} and Pytorch Geometric \cite{pyg} and is trained on two 32-GB Nvidia Titan GPUs. Our code-base is publicly available at \url{https://github.com/MasoudMo/echoglad}.

\subsection{Results}
\label{sec: quant_results}
We evaluate models using Mean Absolute Error (MAE) in mm, and Mean Percent Error (MPE) in percents, which is formulated as $\text{MPE} = 100\times\frac{|L_{\text{pred}} - L_{\text{true}}|}{L_{\text{true}}}$, where $L_{\text{pred}}$ and $ L_{\text{true}}$ are the prediction and ground truth values for every measurement. We also report the Success Detection Rate (SDR) for LVID for 2 and 6 mm thresholds. This rate shows the percentage of samples where the absolute error between ground truth and LVID predictions is below the specific threshold. These thresholds are chosen based on the healthy ranges for IVS (0.6-1.1cm), LVID (2.0-5.6cm), and LVPW (0.6-0.1cm). Hence, the 2 mm threshold provides a stringent evaluation of the models, while the 6 mm threshold facilitates the assessment of out-of-distribution performance.

\textbf{In-Distribution (ID) Quantitative Results.}
In \cref{tab:results1}, we compare the performance of our model with previous works in the ID setting where the training and test sets come from the same distribution (\eg, the same clinical setting), we separately train and test the models on the private and the public dataset. \textit{The results for the public dataset are provided in the supp. material (Table 1).}

\textbf{Out-of-Distribution (OOD) Quantitative Results.}
To investigate the generalization ability of our model compared to previous works, we train all models on the private dataset (which consists of a larger number of samples compared to UIC), and test the trained models on the public UIC dataset as shown in \cref{tab:results3}. 
Based on our visual assessment, the UIC dataset looks very different compared to the private dataset, thus serving as an OOD test-bed. 

\textbf{Qualitative Results.} \textit{Failure cases are shown in supp. material (Fig. 3).}

\begin{table}
% \scriptsize
\caption{\textbf{Quantitative results} on the private test set for models trained on the private training set. We see that our model has the best average performance over the three measurements, which shows the superiority of our model in the in-distribution setting for high-data regime.}
\label{tab:results1}
\begin{center}
\begin{tabular}{l|ccc|ccc|cc}

\multicolumn{1}{c|}{Model} &
  \multicolumn{3}{c|}{MAE {[}mm{]} $\downarrow$} &
  \multicolumn{3}{c|}{MPE {[}\%{]} $\downarrow$} &
  \multicolumn{2}{c}{SDR{[}\%{]} of LVID $<$ $\uparrow$} \\ 
%   \cline{2-7} 
\multicolumn{1}{c|}{} &
  \multicolumn{1}{c|}{LVID} &
  \multicolumn{1}{c|}{IVS} &
  LVPW &
  \multicolumn{1}{c|}{LVID} &
  \multicolumn{1}{c|}{IVS} &
  LVPW &
  \multicolumn{1}{c|}{2.0 mm} &
  6.0 mm \\ \midrule\midrule
Gilbert \etal \cite{gilbert2019automated}&
  \multicolumn{1}{c|}{2.9} &
  \multicolumn{1}{c|}{1.4} &
  1.4 &
  \multicolumn{1}{c|}{6.5} &
  \multicolumn{1}{c|}{14.5} &
  15.2 &
  \multicolumn{1}{c|}{48.1} &
  88.9\\ 
Lin \etal \cite{lin2021reciprocal} &
  \multicolumn{1}{c|}{9.4} &
  \multicolumn{1}{c|}{11.2} &
  9.0 &
  \multicolumn{1}{c|}{21.2} &
  \multicolumn{1}{c|}{116.5} &
  92.9 &
  \multicolumn{1}{c|}{26.0} &
  49.1 \\ 
McCouat \etal \cite{contour} &
    \multicolumn{1}{c|}{\textbf{2.2}} &
    \multicolumn{1}{c|}{1.3} &
    1.4 &
    \multicolumn{1}{c|}{\textbf{4.8}} &
    \multicolumn{1}{c|}{13.5} &
    15.1 &
    \multicolumn{1}{c|}{58.3} &
    93.9  \\ 
Chen \etal \cite{pyramid} &
    \multicolumn{1}{c|}{2.3} &
    \multicolumn{1}{c|}{1.2} &
    1.2 &
    \multicolumn{1}{c|}{5.2} &
    \multicolumn{1}{c|}{12.6} &
    13.8 &
    \multicolumn{1}{c|}{60.4} &
    92.6 \\ 
Duffy \etal \cite{echonetlvh} &
    \multicolumn{1}{c|}{2.5} &
    \multicolumn{1}{c|}{1.2} &
    1.2 &
    \multicolumn{1}{c|}{5.4} &
    \multicolumn{1}{c|}{13.2} &
    13.5 &
    \multicolumn{1}{c|}{52.1} &
    93.0 \\ 
Ours &
\multicolumn{1}{c|}{\textbf{2.2}} &
\multicolumn{1}{c|}{\textbf{1.1}} &
\textbf{1.1} &
\multicolumn{1}{c|}{\textbf{4.8}} &
\multicolumn{1}{c|}{\textbf{11.2}} &
\textbf{12.2} &
    \multicolumn{1}{c|}{\textbf{62.4}} &
    \textbf{94.4}
\\
\end{tabular}
\end{center}
\end{table}

\begin{table}
% \scriptsize
\caption{\textbf{Quantitative results} on the public UIC test set for models trained on the private training set. This table shows the out-of-distribution performance of the models when trained on a larger dataset and tested on a smaller external dataset. We can see that in this case, our model outperforms previous works by a large margin, which attests to the generalizability of our framework.}
\label{tab:results3}
\begin{center}
\begin{tabular}{l|ccc|ccc|cc}

\multicolumn{1}{c|}{Model} &
  \multicolumn{3}{c|}{MAE {[}mm{]} $\downarrow$} &
  \multicolumn{3}{c|}{MPE {[}\%{]} $\downarrow$} &
  \multicolumn{2}{c}{SDR{[}\%{]} of LVID $<$ $\uparrow$} \\ 
%   \cline{2-7} 
\multicolumn{1}{c|}{} &
  \multicolumn{1}{c|}{LVID} &
  \multicolumn{1}{c|}{IVS} &
  LVPW &
  \multicolumn{1}{c|}{LVID} &
  \multicolumn{1}{c|}{IVS} &
  LVPW &
  \multicolumn{1}{c|}{2.0 mm} &
  6.0 mm \\ \midrule\midrule
Gilbert \etal \cite{gilbert2019automated}&
  \multicolumn{1}{c|}{9.5} &
  \multicolumn{1}{c|}{4.8} &
  4.1 &
  \multicolumn{1}{c|}{23.5} &
  \multicolumn{1}{c|}{32.3} &
  26.8 &
\multicolumn{1}{c|}{22.5} &
52.2 \\ 
Lin \etal \cite{lin2021reciprocal} &
  \multicolumn{1}{c|}{51.5} &
  \multicolumn{1}{c|}{51.7} &
  41.3 &
  \multicolumn{1}{c|}{121.0} &
  \multicolumn{1}{c|}{375.8} &
  298.0 &
\multicolumn{1}{c|}{11.3} &
24.6 \\ 
McCouat \etal \cite{contour} &
    \multicolumn{1}{c|}{5.9} &
    \multicolumn{1}{c|}{3.6} &
    4.4 &
    \multicolumn{1}{c|}{18.5} &
    \multicolumn{1}{c|}{30.5} &
    36.4 &
\multicolumn{1}{c|}{34.6} &
72.3 \\ 
Chen \etal \cite{pyramid} &
    \multicolumn{1}{c|}{7.4} &
    \multicolumn{1}{c|}{5.3} &
    6.9 &
    \multicolumn{1}{c|}{22.5} &
    \multicolumn{1}{c|}{49.4} &
    62.4 &
\multicolumn{1}{c|}{28.9} &
65.3 \\ 
Duffy \etal \cite{echonetlvh} &
    \multicolumn{1}{c|}{13.7} &
    \multicolumn{1}{c|}{4.1} &
    5.5 &
    \multicolumn{1}{c|}{36.8} &
    \multicolumn{1}{c|}{36.4} &
    45.4 &
\multicolumn{1}{c|}{6.2} &
20.6 \\ 
Ours &
\multicolumn{1}{c|}{\textbf{5.8}} &
\multicolumn{1}{c|}{\textbf{2.8}} &
\textbf{4.3} &
\multicolumn{1}{c|}{\textbf{18.4}} &
\multicolumn{1}{c|}{\textbf{23.8}} &
\textbf{34.6} &
\multicolumn{1}{c|}{\textbf{35.8}} &
\textbf{\textbf{74.9}}
 \\
\end{tabular}
\end{center}
\end{table}

\begin{table}
  \caption{\textbf{Ablation results} on the validation set of our private dataset. Vanilla U-Net uses a simple U-Net model, while U-Net Main Graph only uses the pixel-level graph (no aux. graphs). Main Model is our proposed approach. Lastly, Single-Scale Loss has the same framework as the Main Model but only computes the loss for the model's predictions on the main graph (no multi-scale loss).}
  \label{tab:ablation1}
\begin{center}
\begin{tabular}{l|ccc}

\multicolumn{1}{c|}{Model} &
  \multicolumn{3}{c}{MPE {[}\%{]}} \\ 
%   \cline{2-7} 
\multicolumn{1}{c|}{} &
  \multicolumn{1}{c|}{LVID} &
  \multicolumn{1}{c|}{IVS} &
  LVPW \\ \midrule\midrule
Vanilla U-Net &
  \multicolumn{1}{c|}{5.31} &
  \multicolumn{1}{c|}{13.17} &
  13.47 \\ 
U-Net Main Graph &
  \multicolumn{1}{c|}{4.98} &
  \multicolumn{1}{c|}{11.67} &
  12.78 \\ 
Single-Scale Loss &
  \multicolumn{1}{c|}{5.41} &
  \multicolumn{1}{c|}{12.37} &
  12.8 \\ 
Main Model &
  \multicolumn{1}{c|}{\textbf{4.91}} &
  \multicolumn{1}{c|}{\textbf{11.45}} &
  \textbf{12.36} \\ 

\end{tabular}
\end{center}
\end{table}

\textbf{Ablation Studies.}
In Table \ref{tab:ablation1}, we show the benefits of a hierarchical graph representation with a multi-scale objective for the task of LV landmark detection. \emph{We provide a qualitative view of the ablation study in supp. material (Fig. 4).}

\section{Conclusion and Future Work}
In this work, we introduce a novel hierarchical GNN for LV landmark detection. The model performs better than the state-of-the-art on most measurements without relying on label smoothing. We attribute this gain in performance to two main contributions. First, our choice of representing each frame with a hierarchical graph has facilitated direct interaction between pixels at differing scales. This approach is effective in capturing the nuanced dependencies amongst the landmarks, bolstering the model's performance. Secondly, the implementation of a multi-scale objective function as a supervisory mechanism has enabled the model to construct a superior inductive bias. This approach allows the model to leverage simpler tasks to optimize its performance in the more challenging pixel-level landmark detection task.

For future work, we believe that the scalability of the framework for higher-resolution images must be studied. Additionally, extension of the model to video data can be considered since the concept of intra-scale and inter-scale edges connecting nodes could be extrapolated to include temporal edges linking similar spatial locations across frames. Such an approach could greatly enhance the model's performance in unlabeled frames, mainly through the enforcement of consistency in predictions from frame to frame. 

\bibliographystyle{splncs04}
\bibliography{ref}

\include{supp}

\end{document}

%% file: supp.tex
\title{Supplementary Material}
\author{Masoud Mokhtari et al.}
\authorrunning{M. Mokhtari et al.}
\institute{Electrical and Computer Engineering, University of British Columbia,
Vancouver, BC, Canada}
\maketitle              % typeset the header of the contribution

\begin{figure}[h!]
    \centering
    \includegraphics[width=0.88\linewidth]{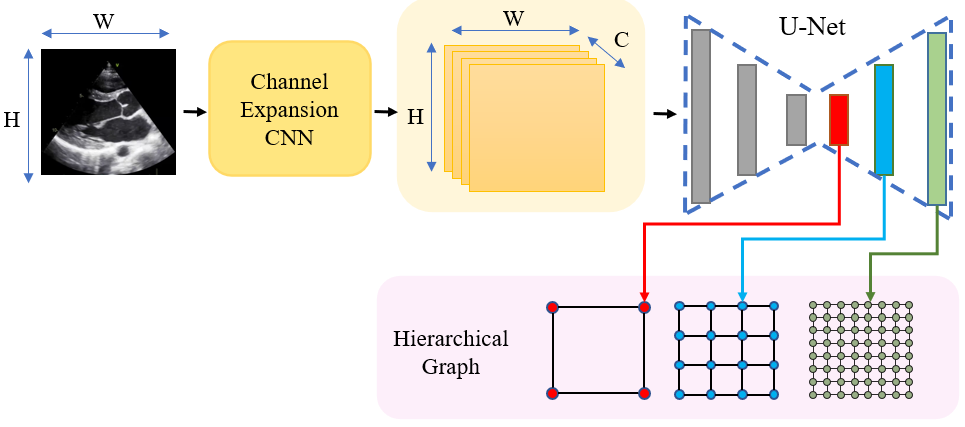}
    \caption{\textbf{Feature Generation for Graph Nodes} - A CNN is initially used to expand the number of the feature maps. The intermediate features of the decoder part of a U-Net are then used as node features such that deeper representations correspond to node features of finer graphs.}
    \label{fig: node_feats}
\end{figure}

\begin{table}[h!]
% \scriptsize
\caption{\textbf{Quantitative results} on the public UIC test set for models trained on the UIC training set. Although the number of training samples is much lower for UIC compared to our private dataset, we see that our model still outperforms previous works on average over the three measurements, which showcases the accuracy of our model in the low-data regime and in-distribution settings. Lin \etal is excluded since they require video inputs.}
\label{tab:results2}
\begin{center}
\begin{tabular}{l|ccc|ccc|cc}

\multicolumn{1}{c|}{Model} &
  \multicolumn{3}{c|}{MAE {[}mm{]} $\downarrow$} &
  \multicolumn{3}{c|}{MPE {[}\%{]} $\downarrow$} &
  \multicolumn{2}{c}{SDR{[}\%{]} of LVID $<$ $\uparrow$} \\ 
%   \cline{2-7} 
\multicolumn{1}{c|}{} &
  \multicolumn{1}{c|}{LVID} &
  \multicolumn{1}{c|}{IVS} &
  LVPW &
  \multicolumn{1}{c|}{LVID} &
  \multicolumn{1}{c|}{IVS} &
  LVPW &
  \multicolumn{1}{c|}{2.0 mm} &
  6.0 mm \\ \midrule\midrule
Gilbert \etal &
  \multicolumn{1}{c|}{5.2} &
  \multicolumn{1}{c|}{2.5} &
  3.1 &
  \multicolumn{1}{c|}{12.2} &
  \multicolumn{1}{c|}{19.0} &
  22.7 &
    \multicolumn{1}{c|}{32.2} &
    70.0 \\ 
McCouat \etal &
    \multicolumn{1}{c|}{2.5} &
    \multicolumn{1}{c|}{1.6} &
    2.4 &
    \multicolumn{1}{c|}{7.5} &
    \multicolumn{1}{c|}{14.8} &
    19.9 &
    \multicolumn{1}{c|}{56.4} &
    91.7 \\ 
Chen \etal &
    \multicolumn{1}{c|}{2.3} &
    \multicolumn{1}{c|}{\textbf{1.5}} &
    2.3 &
    \multicolumn{1}{c|}{7.1} &
    \multicolumn{1}{c|}{\textbf{12.5}} &
    21.4 &
    \multicolumn{1}{c|}{57.3} &
    94.6 \\ 
Yao \etal &
    \multicolumn{1}{c|}{15.4} &
    \multicolumn{1}{c|}{8.8} &
    9.2 &
    \multicolumn{1}{c|}{44.8} &
    \multicolumn{1}{c|}{78.5} &
    80.5 &
    \multicolumn{1}{c|}{7.5} &
    24.6 \\ 
Duffy \etal &
    \multicolumn{1}{c|}{8.7} &
    \multicolumn{1}{c|}{3.4} &
    3.8 &
    \multicolumn{1}{c|}{24.8} &
    \multicolumn{1}{c|}{34.8} &
    34.1 &
    \multicolumn{1}{c|}{13.7} &
    42.4 \\ 
Ours &
\multicolumn{1}{c|}{\textbf{2.2}} &
\multicolumn{1}{c|}{\textbf{1.5}} &
\textbf{1.9} &
\multicolumn{1}{c|}{\textbf{6.2}} &
\multicolumn{1}{c|}{14.0} &
\textbf{16.9} &
    \multicolumn{1}{c|}{\textbf{58.9}} &
    \textbf{94.9}
 \\
\end{tabular}
\end{center}
\end{table}

\begin{figure}[h!]
    \centering
    \settowidth\rotheadsize{Example 1}
    \begin{tabular}{@{\hspace{0.5mm}}c@{\hspace{0.5mm}}c@{\hspace{0.5mm}}c@{\hspace{0.5mm}}c}
        \includegraphics[width=0.22\textwidth,trim={0 0 0 0},clip]{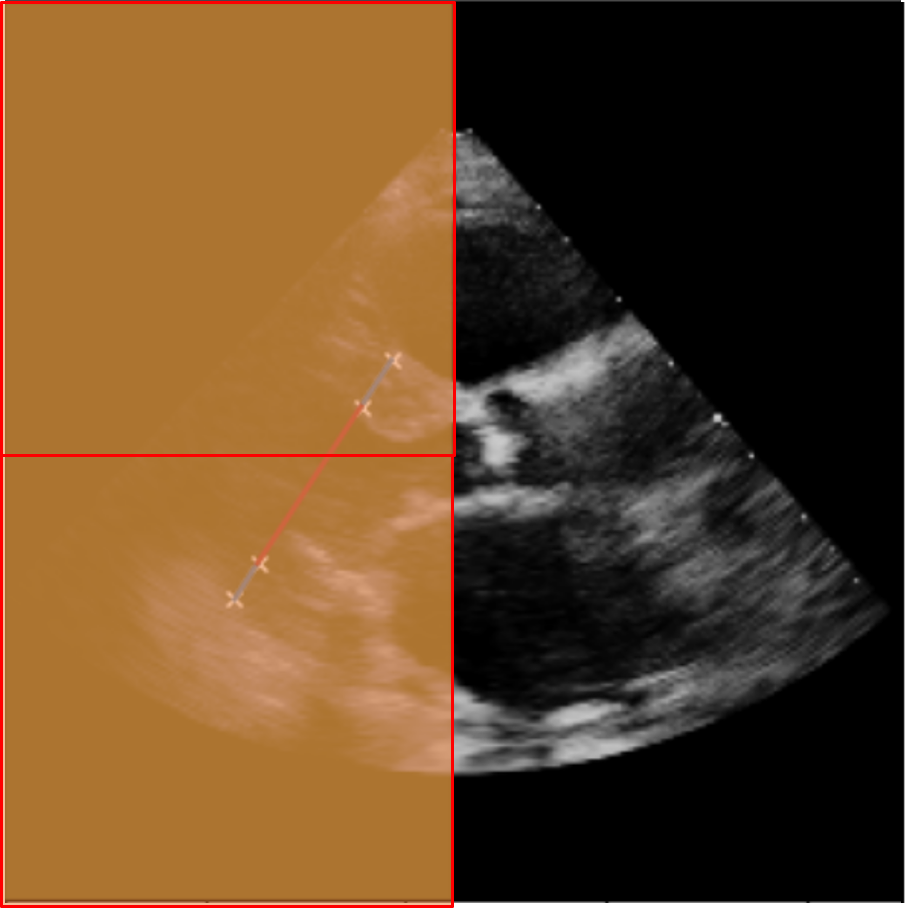} & 
        \includegraphics[width=0.22\textwidth,trim={0 0 0 0},clip]{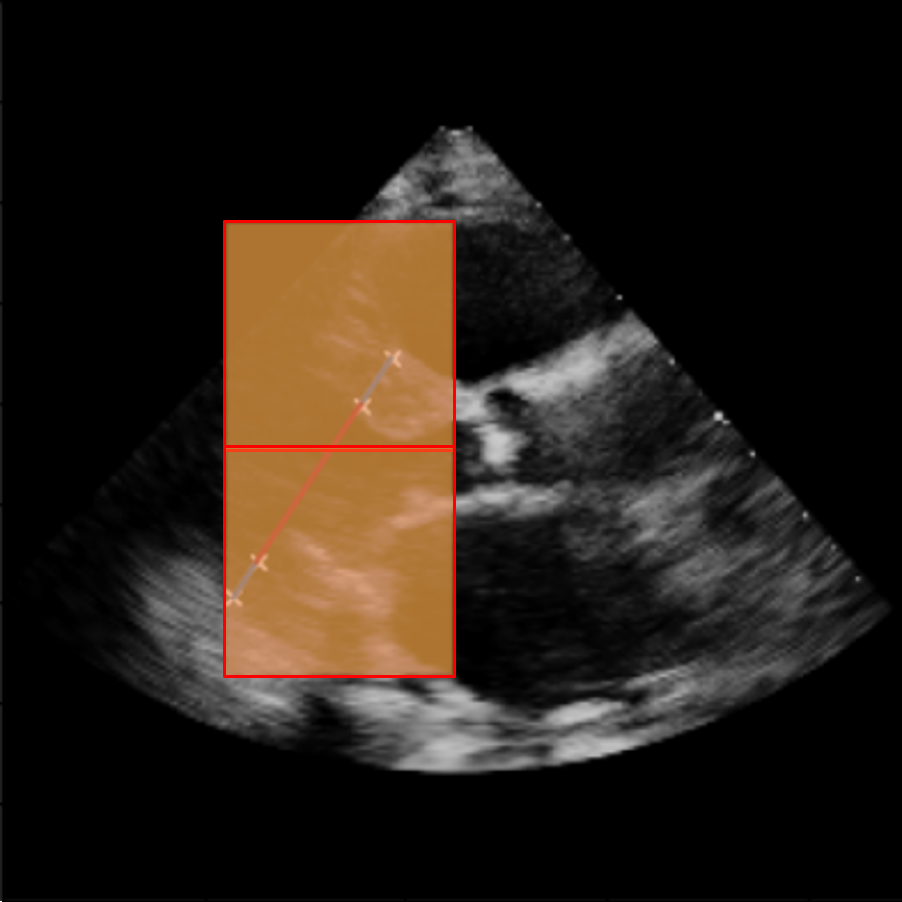} &
        \\
        {\footnotesize } 
        {\footnotesize $112\times112$} 
        & {\footnotesize $56\times56$} 
        \\ \addlinespace[0.5mm]
        \includegraphics[width=0.22\textwidth,trim={0 4 0 0},clip]{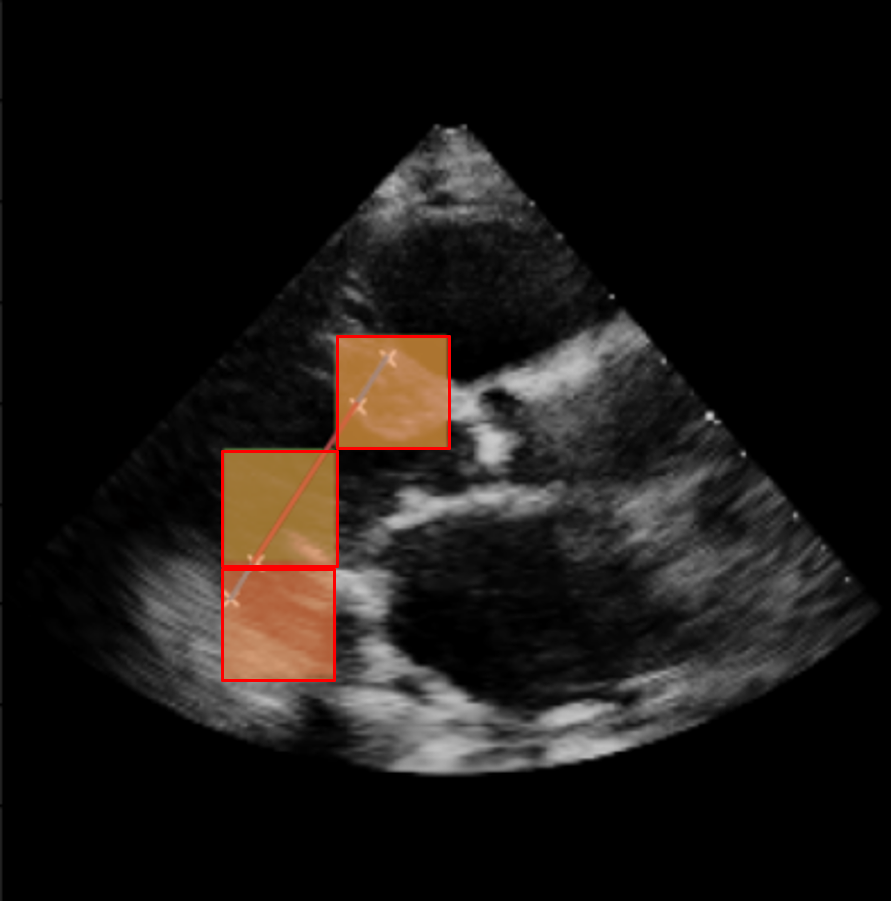} & 
        \includegraphics[width=0.22\textwidth,trim={0 0 0 0},clip]{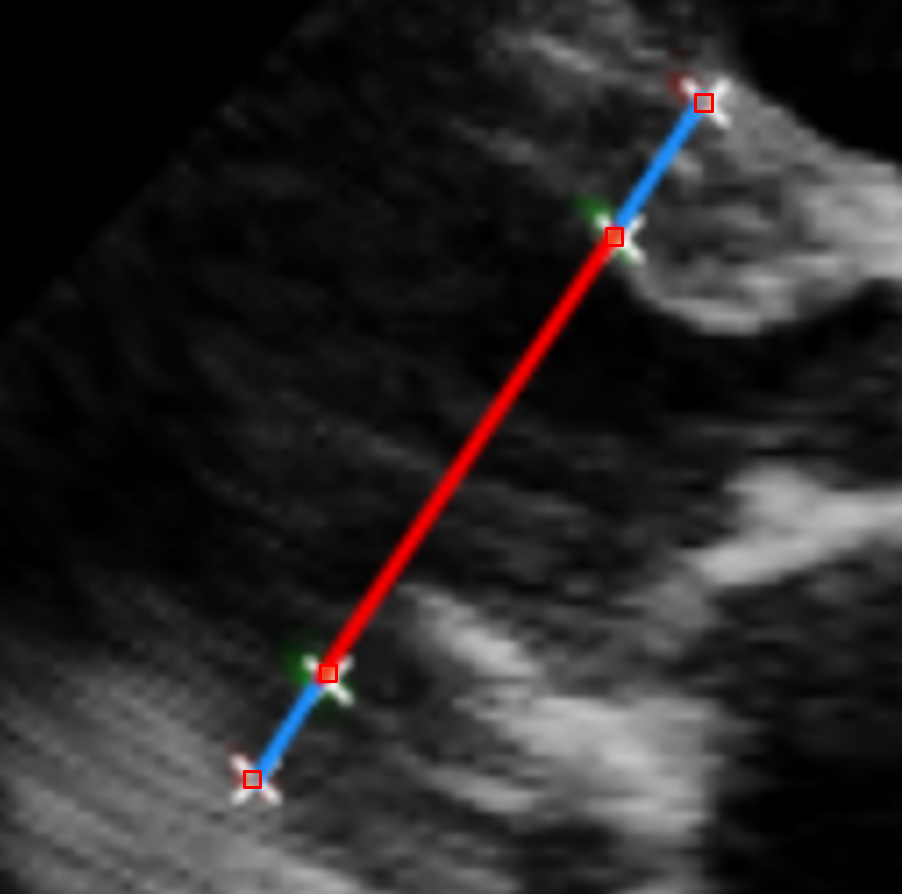} &
        \\
        {\footnotesize } 
        {\footnotesize $28\times28$}
        & {\footnotesize $1\times1$}
    \end{tabular}
    % \vspace{-0.3cm}
    \caption*{Fig. 2: \textbf{Hierarchical predictions} - An example of the model's prediction for an input echo. We show the model's prediction for the case where only three auxiliary graphs are used. We see that the model is learning the LV landmarks on different resolutions to achieve high accuracy for the pixel-level task. We show zoomed-in versions of the higher resolution task to enable comparison.} 
\end{figure}

\begin{figure}[H]
    \centering
            \settowidth\rotheadsize{Example 1}
    \begin{tabular}{@{\hspace{0mm}}c@{\hspace{0.5mm}}c@{\hspace{0.5mm}}c@{\hspace{0.5mm}}c}
        \rothead{\centering Ground Truth} &
        \includegraphics[width=0.22\textwidth,trim={0 0 0 0},clip,valign=m]{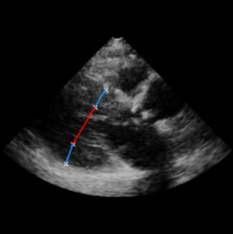} & 
        \includegraphics[width=0.22\textwidth,trim={0 1 0 0},clip,valign=m]{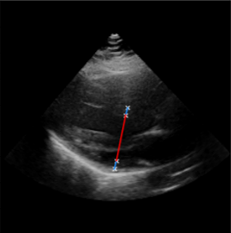} &            
        \\ \addlinespace[0.5mm]
        \rothead{\centering Predictions} &
        \includegraphics[width=0.22\textwidth,trim={0 0 0 0},clip,valign=m]{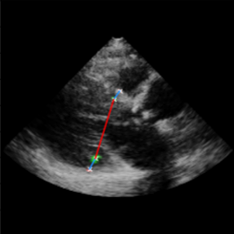} & 
        \includegraphics[width=0.22\textwidth,trim={0 1 0 0},clip,valign=m]{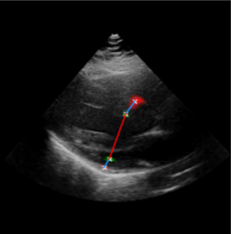} &
        \\ 
        {\footnotesize  } 
        & {\footnotesize Failure 1} 
        & {\footnotesize Failure 2}
        \\
    \end{tabular}
       \caption*{Fig. 3:\textbf{ Qualitative visualization} of our model on two failure cases from the test set of our private dataset. 
    The Failure 1 example is a low-quality PLAX image that also corresponds to a patient with severe LVH, a scenario that happens rarely in our dataset. 
    The Failure 2 example belongs to a case with a low quality of PLAX with unclear boundaries for the walls and the chambers of the LV.}
\end{figure}

\begin{figure}[H]
    \centering
    % \settowidth{\rotheadsize}{\bfseries Long long\quad}
    \settowidth\rotheadsize{Example 1}
    \begin{tabular}{@{\hspace{0mm}}c@{\hspace{0.5mm}}c@{\hspace{0.5mm}}c@{\hspace{0.5mm}}c@{\hspace{0.5mm}}c@{\hspace{0.5mm}}c@{\hspace{0.5mm}}c@{\hspace{0.5mm}}c@{\hspace{0.5mm}}c@{\hspace{0.5mm}}c}
        \rothead{\centering Example 1} &
        \includegraphics[width=0.188\textwidth,trim={0 0 0 0},clip,valign=m]{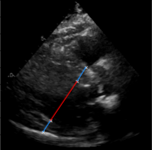} & 
        \includegraphics[width=0.188\textwidth,trim={0 0 0 0},clip,valign=m]{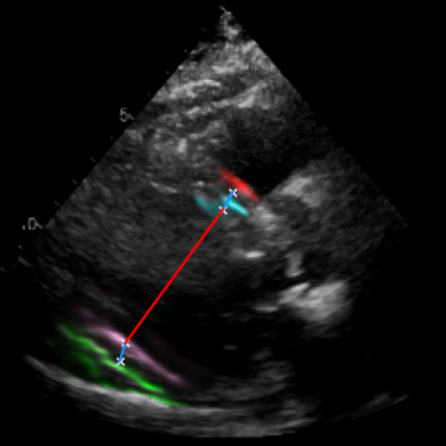} &         
        \includegraphics[width=0.188\textwidth,trim={0 0 0 0},clip,valign=m]{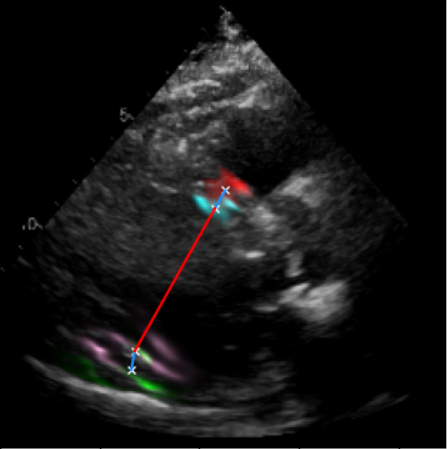} & 
        \includegraphics[width=0.188\textwidth,trim={0 0 0 0},clip,valign=m]{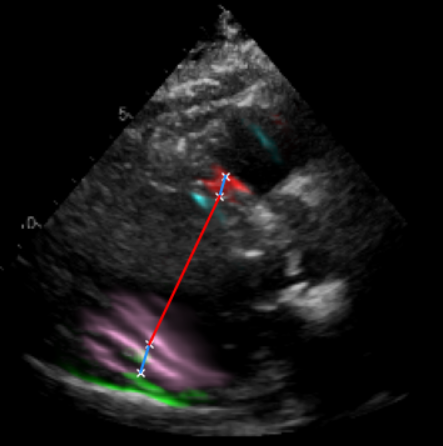} &   
        \includegraphics[width=0.188\textwidth,trim={0 0 0 0},clip,valign=m]{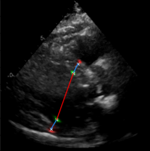} &   
        \\ \addlinespace[1mm]
        \rothead{\centering Example 2} &
        \includegraphics[width=0.188\textwidth,trim={0 0 0 0},clip,valign=m]{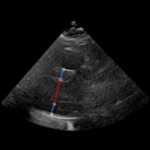} & 
        \includegraphics[width=0.188\textwidth,trim={0 0 0 0},clip,valign=m]{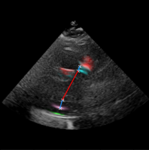} &         
        \includegraphics[width=0.188\textwidth,trim={0 0 0 0},clip,valign=m]{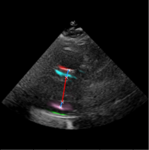} & 
        \includegraphics[width=0.188\textwidth,trim={0 0 0 0},clip,valign=m]{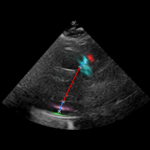} &     
        \includegraphics[width=0.188\textwidth,trim={0 0 0 0},clip,valign=m]{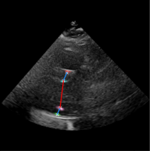} &   
        \\ 
        {\footnotesize  } 
        & {\footnotesize Ground Truth} 
        & {\footnotesize V. U-Net} 
        & {\footnotesize U. M. Graph}
        & {\footnotesize SSL}
        & {\footnotesize Main Model}
        \\
    \end{tabular}
    \caption*{Fig. 4: \textbf{Qualitative ablation results} for the model architecture. Landmark heatmaps from top to bottom are color-coded with red, cyan, pink and green, respectively. We see that Vanilla U-Net (V. U-Net) struggles to make confident and accurate landmark predictions. While the addition of a main grid graph in U-Net Main Graph (U. M. Graph) relatively increases model's performance, it still does not produce accurate results. In contrast, the Main Model produces confident prediction heatmaps by relying on a hierarchical graph representation as well as multi-scale objectives. We also see that the removal of the multi-scale objective (Single-Scale Loss (SSL)) degrades performance.} 
    \label{fig: ablation_qual} 
\end{figure}
\null
\vfill

% \end{document}